\pdfoutput=1

\documentclass[11pt]{article}

\usepackage{acl}

\usepackage{times}
\usepackage{latexsym}

\usepackage[T1]{fontenc}

\usepackage[utf8]{inputenc}
\usepackage{microtype}

\usepackage{times}
\usepackage{latexsym}
\usepackage{amsmath}
\usepackage{graphicx}
\usepackage{subfigure}
\usepackage{bbm}
\usepackage{multirow}
\usepackage{enumerate}
\usepackage{algorithm}
\usepackage[noend]{algpseudocode}
\usepackage{xcolor}
\usepackage{float}

\usepackage{amsmath, amssymb, amsthm, amscd, amsfonts, booktabs}
\usepackage{IEEEtrantools}
\usepackage{enumitem}
\usepackage[OT1]{fontenc}
\usepackage{soul}

\definecolor{RED}{rgb}{1,0,0}

\newtheorem{theorem}{Theorem}

\newtheorem{lemma}[theorem]{Lemma}

\theoremstyle{definition}

\theoremstyle{remark}

\newcommand*{\affaddr}[1]{#1} 
\newcommand*{\affmark}[1][*]{\textsuperscript{#1}}
\newcommand*{\email}[1]{\texttt{#1}}

\title{Few Clean Instances Help Denoising Distant Supervision}

\author{%
Yufang Liu$^*$\affmark[1], Ziyin Huang$^*$\affmark[1], Yijun Wang\affmark[2],Changzhi Sun\affmark[3], \\
\textbf{Man Lan\affmark[1], Yuanbin Wu\affmark[1], Xiaofeng Mou\affmark[4] and Ding Wang\affmark[4]}\\
\affaddr{\affmark[1]School of Computer Science and Technology, East China Normal University}\\
\affaddr{\affmark[2]Department of Computer Science and Engineering, Shanghai Jiao Tong University}\\
\affaddr{\affmark[3]Bytedance AI Lab, \affmark[4]AI Innovation Center, Midea Group}\\
\email{\{yfliu.antlp, zyhuang.cs\}@gmail.com, ybwu@cs.ecnu.edu.cn}\\
}

\begin{document}
\maketitle
\def\thefootnote{*}\footnotetext{These authors contributed equally.}\def\thefootnote{\arabic{footnote}}  

\begin{abstract}
    Existing distantly supervised relation extractors usually
    rely on noisy data for both model training and evaluation,
    which may lead to garbage-in-garbage-out systems.
    To alleviate the problem,
    we study whether a small clean
    dataset could help improve the quality of distantly supervised models.
    We show that besides getting a more convincing evaluation of models,
    a small clean dataset also helps us to build more robust denoising models.
    Specifically,
    we propose a new criterion for clean instance selection based on influence
    functions. It collects sample-level evidence for recognizing good instances
    (which is more informative than loss-level evidence).
    We also propose a teacher-student mechanism
    for controlling purity of intermediate results when bootstrapping the clean set.
    The whole approach is model-agnostic and demonstrates strong performances
    on both denoising real (NYT) and synthetic noisy datasets.\footnote{
    Our codes are publicly available at: \url{https://github.com/Airuibadi/IF_DSRE}.}
\end{abstract}

\section{Introduction}




 

Distant supervision was introduced to tackle 
the lacking training data problem
in information extraction tasks \cite{mintz-etal-2009-distant}.
By aligning relation triples in knowledge bases (KB) with free texts,
it automatically builds labelled sentence instances and 
easily extends the scale of training set to hundreds of thousands samples.
Due to this great scalability, distantly supervised entity relation extractors
have been extensively studied in the past decade.

Like other weak signals, 
the major problem about these automatically generated datasets is label noise:
not all aligned sentences carry the same semantic of a KB triple
(e.g., not all sentences containing ``Obama'' and ``United States'' 
express a ``born in'' relation).
Some applications 
(e.g., slot filling of the TAC KBP track \cite{ji-grishman-2011-knowledge})
could be less affected with the help of \emph{instance bags},
which only needs to seek one correct instance among a bag of aligned sentences.
For a more general setting which aims to correctly detect relations on 
\emph{individual sentences} 
\cite{miwa-bansal-2016-end,sun-etal-2018-extracting,
wadden-etal-2019-entity,wang-etal-2020-pre},
however, the noisy labels make both 
learning and evaluation of models vulnerable:
we may draw a flawed conclusion by using a dirty test set for
a model learned with a dirty training set.


Many methods have been proposed to reduce noise labels (\emph{denoise}) 
in distant supervision.
For bag-level applications, models often 
rely on attention scores to either filter bad instances
inside a bag (intra-bag attentions, \citealt{lin-etal-2016-neural}) 
or filter bags full of noisy instances 
(inter-bag attentions, \citealt{ye-ling-2019-distant}).
The dilemma there is that, 
while we expect attention scores to indicate correct labels, 
we have to train them to fit noisy labels 
since ground truth labels are noisy.
The same difficulty also exists in recent instance-level denoising methods
\cite{qin-etal-2018-dsgan,qin-etal-2018-robust}
where the reward of denoising an instance is obtained by querying noisy labels.
Therefore, not only the extraction models but also denoising models
are questionable if only noisy labels are given.


In this paper, we would like to restate the importance of 
trustful data (\emph{clean dataset}) 
in building large-scale information extraction systems.
Specifically, if a \emph{small} clean dataset ($\approx 10^2$ samples) is available, 
we ask whether the robustness of 
both the denoising model and final extraction model could be improved.

We start from training a relation classifier on the clean set
and propose a new criterion to select good instances from the dirty set.
The main idea is that
if a testing instance is correctly labelled by distant supervision,
some instances in the clean set should support it,
and if we remove those support instances, 
prediction error of the testing instance will increase.
Comparing with previous work, the criterion 
is based on perturbation analyses of classifiers instead of 
directly using output probabilities (scores) of classifiers.
Our tool is \emph{influence function} (IF; \citealt{cook1982,koh2017understanding})
which can effectively approximate how a classifier's parameters change
when removing a training point.

Next, to incrementally explore the dirty set,
we compile our instance selection algorithm into a bootstrapping process: 
training a classifier on the current clean set,
selecting new clean instances using the classifier 
and retraining the classifier on the updated clean set.
The key challenge is 
how to control purity of those intermediate datasets:
one noisy instance may bring more noisy instances.
Existing works are either lack of such strategy,
or use heuristic thresholds on classifiers or dataset size
\cite{jia-etal-2019-arnor}.
Here, we propose a \emph{teacher-student style} update
for learning intermediate classifiers.
It gradually controls the distance between the current model and history models
by regularizing discrepancy of their predictions.

Our whole system could be deemed as a data preprocessing method.
Comparing with in-model denoising method (e.g., attention scores),
it outputs a new clean set which can be applied to any 
information extraction models (\emph{model-agnostic}).
We conduct experiments on both real distantly supervised datasets (NYT)
and synthetic datasets (built on ACE05).
The results demonstrate that besides effectively selecting good instances,
the influence-function-based criterion
can stratify noisy instances according their difficulties for prediction
(or importances for a better extractor).
We also find that the teacher-student update especially helps when the proportion
of incorrectly labelled instances is large.
Finally, when learned with clean sets built by our methods,
we are able to achieve competitive extraction performances
on \emph{manually} labelled testing set.

\section{Preliminary}
\label{sec:pre}

\paragraph{Distantly Supervised Relation Classification}

Given an entity pair $(e_h, e_t)$ and a sentence $s$ containing the pair,
we consider the task of determining whether the entity pair expresses
certain relation $r \in R$, where $R$ is the set of relation types
($\mathtt{None}$ indicates no relation).
Denote $x=(s, e_h, e_t, r)$ to be an \emph{instance}, 
$y\in\{0, 1\}$ to indicate whether $x$ is positive or negative,
and $D=\{(x_i, y_i)\}_{i=1}^{|D|}$ to be a set of labelled instances.
For simplicity, we also define $z=(x, y)$.

In the distant supervision setting, 
instances in $D$ are automatically obtained
by aligning plain text and knowledge bases:
for a KB triple $(e_h, e_t, r)$, every sentence containing
$(e_h, e_t)$ is labelled with $r$.
Obviously, $D$ is a dirty set
with both false positives (sentences don't match the semantic of $r$) 
and false negatives (sentences expressing relation $r$ while been labelled with $\mathtt{None}$ 
due to incompleteness of KB).
Here, we focus on false positives (much more serious in current datasets)
and 
don't consider false negatives for its very low quantity.
The denoising task is thus to find 
$D' \subset D$ containing correctly labelled
instances (especially, positive instances).

\paragraph{Influence Function}
\cite{cook1982,koh2017understanding} provides a
way to estimate how individual training instances influence a model.
Typically, for a testing instance $(x', y')$, 
it efficiently answers the question that
if a training instance $(x, y)$ is removed 
how the model's prediction on $(x', y')$ changes.

Denote $\mathcal{L}(z, \theta)$ to be 
a convex loss function of $z$ with parameter $\theta$, and 
$\hat{\theta} \triangleq 
{\arg \min}_\theta \frac{1}{n}\sum_{i=1}^n \mathcal{L}(z_i,\theta)$
to be the optimal model parameter learned on a training set ($n$ is the set size). 
To study a training instance $z$'s influence on $\hat{\theta}$, 
influence function considers an $\epsilon$ up-weight on $z$.
Define
$\hat{\theta}_{\epsilon, z} \triangleq 
{\arg\min}_{\theta}\frac{1}{n}\sum_{i=1}^{n}
\mathcal{L}(z_i, \theta)+\epsilon \mathcal{L}(z,\theta)$.
Therefore, when $\epsilon = -\frac{1}{n}$,
$\hat{\theta}_{\epsilon, z}$ is the new model parameter 
after removing $z$ from the training set.

The key idea of influence function
is that, 
when $\epsilon$ is small (or training set size $n$ is large),
with the first order Taylor approximation,
we can measure the difference between $\hat{\theta}$
and $\hat{\theta}_{\epsilon,z}$ 
without retraining the model,
\begin{IEEEeqnarray*}{c}
\hat{\theta}_{\epsilon,z}-\hat{\theta} 
\approx -\epsilon H_{\hat{\theta}}^{-1}\nabla_{\theta}\mathcal{L}(z,\hat{\theta})
\triangleq
\epsilon \mathcal{I}_{up,params}(z),
\end{IEEEeqnarray*}
where $H_{\theta} = \frac{1}{n}\sum_{i=1}^{n} \nabla^2 \mathcal{L}(z_i, \theta)$ \
is the Hessian matrix of the original loss function.
\footnote{
We follow \cite{koh2017understanding} using a stochastic 
estimation of $H_{\theta}^{-1}$ when computing influence functions.
}

We can also get the change of the model's prediction on 
a testing instance $z'$ by the chain rule,
\begin{IEEEeqnarray*}{rl}
    & \mathcal{L}(z', \hat{\theta}_{\epsilon,z}) - \mathcal{L}(z', \hat{\theta})\\
   \! \approx \! & -\epsilon \nabla_{\theta}\mathcal{L}(z',\hat{\theta})^T
    H_{\hat{\theta}}^{-1}\nabla_{\theta}\mathcal{L}(z,\hat{\theta})
\triangleq
\epsilon \mathcal{I}_{up,loss}(z, z').
\end{IEEEeqnarray*}
We say $z$ \emph{supports} (or is \emph{helpful} to) $z'$ 
if removing $z$ increases the testing loss of $z'$,
that is $\mathcal{S}(z, z')>0$, where
\begin{IEEEeqnarray}{c}
    \label{eq:support_score}
    \mathcal{S}(z, z') \triangleq -\frac{1}{n}\mathcal{I}_{up,loss}(z, z').
\end{IEEEeqnarray}
We will see in the next section that
the group of supporting instances
is a key part in our denoising algorithms.


\section{Utilities of Clean Sets}

We study the task of picking out correct instances ($D'$)
from a distantly supervised dataset $D$ (the dirty set).
As discussed above,
it is not easy for the denoising model to
either correctly evaluate its results
or receive the right learning signals if we only know noisy labels in $D$.
Therefore, departing from previous works,
we additionally require a small clean set $C$ 
($C\cap D=\emptyset$, $|D|\gg|C|$)
which contains trustful annotations of instances (e.g., manually labelled).
In our experiments, $|C|=10$ is enough to bring significant improvement.

We build our denoising model based on a binary classifier $\hat{\theta}$,
which aims to recognize truly labelled instances from $D$.\footnote{
It is also possible to denoise by directly comparing 
similarities among instances (e.g., using patterns or sentence embeddings). 
While these methods are important,
we mainly focus on classifier-based models whose settings
are more analogous to semi-supervised learning or active learning.
Comparing with them is beyond the scope of this paper.
}
The classifier could be learned on $D$ or $C$.
For example, for an instance $z$, 
to test whether it is correctly labelled or not,
a broadly applied principle 
is to query the classifier's confidence on 
predicting $z$'s label:
the lower loss $\mathcal{L}(z, \hat{\theta})$,
the more possible $z$ being correctly labelled.

Here, 
we go one step deeper: besides looking at the loss function,
we could first seek
high impact training samples on the classifier's 
drawing of $\mathcal{L}(z, \hat{\theta})$,
and then collect evidence from them.
For example,
the more clean instances support $z$, 
the more possible $z$ being correctly labelled.
We are going to demonstrate that
by probing the black-box classification process,
we could build more explainable (yet effective)
criteria for selecting instances.

First, from the computation of influence function, 
we can obtain a training instance's influence
on a testing instances (Equation \ref{eq:support_score}).
Then, for a instance $z_d\in D$ 
(as discussed above, we mainly focus on positive $z_d=(x, y)$ where $y=1$), 
we have two possible directions to derive a selection criterion.
\begin{itemize}[leftmargin=*]
    \item \textbf{Criterion 1}.
        We can train a classifier on $D$. 
        $z_d$ is correctly labelled
        if it supports $\hat{\theta}$'s prediction on the clean set $C$.
        Concretely, define 
        $\mathcal{S}(z_d, \star)\triangleq\frac{1}{|C|}\sum_{z_c\in C}\mathcal{S}(z_d, z_c)$
        to be the marginal $\mathcal{S}$ over the testing set,
        \begin{IEEEeqnarray}{c}
            \mathcal{S}(z_d, \star) > 0 ~ ~ \Longrightarrow ~ ~ z_d ~\text{ is correct}.
            \label{eq:c1}
        \end{IEEEeqnarray}
    \item \textbf{Criterion 2}.
        We can also train a classifier on $C$. 
        In this case, $z_d$ is correctly labelled
        if $\hat{\theta}$'s prediction on $z_d$ is supported by the instances in $C$.
        Define $\mathcal{S}(\diamond, z_d)\triangleq
        \frac{1}{|C|}\sum_{z_c\in C}\mathcal{S}(z_c, z_d)$ 
        to be the marginal $\mathcal{S}$ over the training set,
        \begin{IEEEeqnarray}{c}
            \mathcal{S}(\diamond, z_d) > 0 ~ ~ \Longrightarrow ~ ~ z_d ~\text{ is correct}.
            \label{eq:c2}
        \end{IEEEeqnarray}
\end{itemize}


Given the limited budget of clean instances $C$,
the two methods are different in their way of using them.
When taking $C$ as the testing set (Criterion 1),
we emphasize a valid feedback signal for evaluating the denoising model.
On the other hand,
when taking $C$ as the training set (Criterion 2),
we emphasize a clean learning signal for building the denoising model.
We would like to discuss more on their pros and cons.\footnote{
Similarly, we can also select the wrongly labelled instances 
by selecting the lowest influence function scores,
we try to flip the labels and add them to the training set,
but we find it barely working. The possible reason can be that
these instances are positive instances for other entities or relations 
which adds too much noise for our classifier.}

\paragraph{For Criterion 1,}
as $D$ is usually large enough,
we could obtain a sufficiently learned classifier for denoising.
More importantly, a large training set makes
the estimation of influence function more reliable
(Taylor expansion works on small $\epsilon$).
However,
a good fitting of the dirty set could be a double-edged sword,
especially when the proportion of wrongly labelled instances is large:
we do get the influence function estimation right but it may not be applicable 
to our goal of denoising.
We can first consider an ideal setting where
all instances in $D$ are true.
In this case, Equation \ref{eq:c1} is trustable since
the ideal parameter $\hat{\theta}'$ is trustable, and 
it encodes the right information for detecting supporting relationship
between the training and testing set.
However, if a large part of $D$ is false, the classifier $\hat{\theta}$
can diverge from the ideal $\hat{\theta}'$ severely, thus
makes Equation \ref{eq:c1} no longer true
(e.g., a negative $z_d$ could also
satisfy the criterion as $\hat{\theta}$ is learned with noise).

Furthermore, we can have the following characterization of
$|\mathcal{L}(z_d, \hat{\theta}') - \mathcal{L}(z_d, \hat{\theta})|$
if $L(z, \hat{\theta})$ is in the form of log-likelihood,
\begin{IEEEeqnarray*}{c}
    L(z, \hat{\theta}) \! = \! - \log p(y |x, \hat{\theta}) \! = \! 
    -\log \frac{\exp\left( \hat{w}_y^\intercal h(x, \hat{\varphi}) \right)}{Z}
\end{IEEEeqnarray*}
where $\hat{\theta} = [\hat{w}_0, \hat{w}_1, \hat{\varphi}]$, $\hat{w}_0, \hat{w}_1$ are 
class label embedding, $h(x, \hat{\varphi})$ is a learned representation of $x$ (\emph{encoder}),
and $Z$ is the normalizer.

\begin{lemma}
    Let $z=(x, y)\in D$, $z'=(x, y')$ be a relabelled $z$,
    and $\hat{\theta}_{z, z'}$ be the optimal model parameter
    after replacing $z$ with $z'$. 
    Denote $\tau_{x}$ to be the smallest singular value of 
    $\nabla_{\varphi} h(x, \hat{\varphi})$.
    Then for any $z_d\in D$, up to $o(n^{-1})$,
    $|\mathcal{L}(z_d, \hat{\theta}_{z,z'}) - \mathcal{L}(z_d, \hat{\theta})|$
    is lower bounded by
    \begin{IEEEeqnarray*}{c}
        \frac{c}{n}\left( \|h(x, \hat{\varphi})\| + \tau_x\|\hat{w}_y - \hat{w}_{y'}\| \right),
    \end{IEEEeqnarray*}
    for some constant $c$. Proof is in Appendix \ref{proofSection}.
    \label{lemma:lowbound}
\end{lemma}

Therefore, if the classifier $\hat{\theta}$ fits well on 
the dirty set (in the sense of a large $\|\hat{w}_0 - \hat{w}_1\|$), 
$\mathcal{S}(z_d, z_c)$ calculated with $\hat{\theta}$ could be far away from 
its value being calculated with a clean training set (i.e., with $\hat{\theta}'$).
For the case of multiple updates,
since the group version of influence function may not faithfully reflect
the change of parameters \cite{koh2019influence}, 
we are not able to obtain 
similar results with Lemma \ref{lemma:lowbound}. However, our empirical evaluations will show
that performances of Criterion 1 is highly related to the proportion of clean instances in $D$.


\paragraph{For Criterion 2,}
comparing with training with dirty $D$, $C$ contains trustful data, 
thus the implication relation in Equation \ref{eq:c2} is clear after training on $C$.
However, since the clean set is usually small, 
Criterion 2 takes the risk of under-fitting, which makes the prediction on 
$z_d \in D$ not sufficiently exploit structures of clean samples in $C$.
Moreover, the estimation of influence function also becomes unstable on small datasets
(i.e., $\varepsilon$ is larger).
In summary, instead of measuring a wrong $\mathcal{S}(z_d, z_c)$ with good accuracy (like Criterion 1),
Criterion 2 may struggle with measuring the right $\mathcal{S}(z_c, z_d)$ with poor accuracy.

In the following section, we investigate bootstrapping methods to enlarge $C$ incrementally.
We hope that when the number of clean instances becomes larger, we could alleviate 
both under-fitting and poor estimation of influence function gradually.

\section{Bootstrapping the Clean Set}



Given a initial small clean set $C_0$ and a dirty set $D_0$,\footnote{
We use the subscript $t$ to indicate the number of iterations.
In some cases, we drop it for simplicity.}
our bootstrapping framework incrementally updates a denoising classifier $\hat{\theta}$.
At iteration $t$, we first collect a fixed-size clean set $\tilde{C}$ by sampling from $C_t$. 
Second, a denoising classifier $\hat{\theta}$ is trained on the sampled set $\tilde{C}$,
from which we can use influence-function-based scores (Equation \ref{eq:support_score}) to evaluate each instance in $D_t$
and choosing new clean instances $D^c$ from $D_t$.
Third, we update $C_t$ and $D_t$ by merging and excluding instances
in $D^c$ and retraining the denoising model again.
As discussed above, how to control the purity of those 
intermediate clean sets is important (otherwise, we will face the
same challenge as Criterion 1).
We propose teacher-student style update for learning intermediate classifiers. 
It gradually controls the distance between the current model and history models by regularizing discrepancy of their predictions.
We summarize the whole process in Algorithm ~\ref{alg:alg}.
It is worth noting that the output of the bootstrapping process
is a new clean set, on which we could build any 
relation classifier (i.e., \emph{model-agnostic} denoising).

\paragraph{Denoising Classifier}
Our denoising model is a binary classification model.
For each $x=(s, e_h, e_t, r)$, it predicts $y\in\{0,1\}$.
Here, we simply apply a softmax layer on a CNN encoder 
(the same setting of Lemma \ref{lemma:lowbound}).\footnote{
The model could be any existing relation model. For simplicity, we select a simple CNN. }
Specifically,
$h(x, \varphi) = \text{CNN}(\mathbf{s}, \mathbf{p})$,
where $\mathbf{s}$ contains embeddings of words in sentence $s$,
and $\mathbf{p}$ contains position embeddings
which indicates two entities $e_h, e_t$ in the sentence \cite{zeng-etal-2014-relation}.

\paragraph{Sampling}
To obtain a fixed-size clean set $\tilde{C}$,
we randomly sample instances from $C_t$ with replacement.
We keep $|\tilde{C}|=200$ so that the influence function calculation is more efficient.

\paragraph{Fitting}
To fit the relation classifier on the sampled set $\tilde{C}$,
our objective is to minimize
\begin{align}\label{Fit}
    \hat{\theta} = \arg \min_{{\theta}} \sum_{z \in \tilde{C}}
    \mathcal{L}(z, \theta)
\end{align}
The parameters $\hat{\theta}$ are applied in calculating IF.

\paragraph{Evaluating}
After obtaining the parameters $\hat{\theta}$,
to evaluate each instance $ z_d \in D_t$,
we define a score function by Criterion 2 as follows,
\begin{IEEEeqnarray}{c}
    \mathcal{S}(\diamond, z_d) \triangleq\frac{1}{|\tilde{C}|}\sum_{z_c\in \tilde{C}}\mathcal{S}(z_d, z_c)
    \label{calIF}
\end{IEEEeqnarray}
The score $\mathcal{S}(\diamond, z_d)$ is the average 
of clean training instances' influence on the test instance $z_d$.

\paragraph{Selecting}
After obtaining the score for each instance $\mathcal{S}(\diamond, z_d)$,
we can select the clean instances from $D_t$ according $\mathcal{S}(\diamond, z_d) > 0$ (Criterion 2).
In practice,  we observe that adding a relaxation factor works better.
Formally, we denote it as follows,
\begin{equation}\label{selecting1}
\tilde{D}^c_t =\left \{z_d \in D_t |\mathcal{S}(\diamond, z_d) + r > 0\right \}
\end{equation}
where $r$ is a positive number.
In addition, we adopt a majority voting strategy:
we consider not only the current iteration, but also the previous iterations to build the current cleaned set $D^c$, denoted as:
\begin{equation}\label{selecting2}
{D}^c =\left\{z_d | \sum_{i = 0} ^ t \boldsymbol{1} (z_d \in \tilde{D}_{i}^c)  > k\right\}
\end{equation}
where $\boldsymbol{1}(\cdot)$ is the indicator function 
and $k$ is a hyper-parameter.

\paragraph{Updating}
Once we have the set $D^c$, we can update $C_t$ and $D_t$ with simple set operations, denoted as:
\begin{equation}\label{update}
    \begin{split}
    C_{t+1} = C_t \cup {D}^c, ~ ~ ~     D_{t+1} = D_t \setminus {D}^c 
\end{split}
\end{equation}

\begin{algorithm}[t] 
	\caption{Bootstrapping Framework} 
	\begin{algorithmic}[1] 
		\Require  
        \ \ ${C}_0$, ${D}_0, t_\mathrm{max}, k$
		\Ensure
        $D^r$
        \State For the student model, initialize $\theta$ randomly 
        \State For the teacher model, initialize $\bar{\theta}$ with $\theta$
        \For{$t = 0: t_\mathrm{max}$} 
        \State Sample $\tilde{C}$ from $C_t$ randomly
        \State Fit $\theta$  on $\tilde{C}$ by Eq.\ref{new_Fit}
        \State Fit $\bar{\theta}$ by Eq. \ref{EMA}
        \State Evaluate $\mathcal{S}(\diamond, z_d)$   $\forall z_d \in D_t$ by Eq. \ref{calIF}
        \State Select $D^c$ by Eq. \ref{selecting1} and Eq. \ref{selecting2}
        \State  Update $C_t, D_t$ by Eq. \ref{update}
        \EndFor
        \State $D^r = {C}_{t_\mathrm{max}} \setminus {C}_{0}$
	\end{algorithmic} 
	\label{alg:alg}
\end{algorithm}    
    
\paragraph{Teacher-student Mechanism}
Even though we have used an implicit majority voting 
to keep selected instances clean, 
affected by under-fitting and unstable estimation of influence function, 
error instances will inevitably enter the clean set.
Considering our algorithm is based on bootstrapping, 
errors in previous rounds would 
have continuous impact on subsequent selection.
As mentioned before,
the model parameter $\theta$ is easily disturbed by wrong-label instances,
with the error propagation, the $\theta$ would be rotten quickly.  
To avoid this case, we introduce a teacher-student mechanism \cite{DBLP:conf/nips/TarvainenV17}. 

Here,we deem $\theta$ as the student model,
and use another set of model parameters $\bar{\theta}$ as the teacher model,
In the fitting step, we add a consistency regularizer to Equation \ref{Fit}: 
\begin{equation}\label{new_Fit}
    \hat{\theta} = \arg \min_{{\theta}} L(\tilde{C}, \theta) + \alpha \mathrm{KL}(q(* ; \bar{\theta})||p(* ; \theta))
\end{equation}
where the $q$ and $p$ are outputs of teacher model and student model respectively,
the KL-divergence provides a consistency loss and $\alpha$ is a hyper-parameter. 
Furthermore,
the $\bar{\theta}$ would not be updated in Equation \ref{new_Fit},
we update it by exponentially moving average as commonly used in teacher-student method:
\begin{equation}\label{EMA}
    \bar{\theta_t} = \beta \bar{\theta}_{t-1} + (1-\beta)\hat{\theta_{t}}
\end{equation}
Teacher-student mechanism is seen as a regularization term of $\tilde{\theta}$ during fitting,
and we neglect this term when calculating the influence function.
After $t_\mathrm{max}$ times loop, we remove the seed set $C_{0}$ from the $C_\mathrm{max}$ and obtain
our final result $D^r = {C}_{t_\mathrm{max}} \setminus {C}_{0}$.
Then, we could train any model on ${D}^{r}$. 

\section{Experiment}

\subsection{Configurations}
\paragraph{NYT} The NYT dataset is a widely-used distant supervision benchmark,
which is built by \citet{10.1007/978-3-642-15939-8_10} and rearranged 
by \citet{jia-etal-2019-arnor}.
The training set is annotated with distant supervision while both development set and test set are manually annotated.
For this dataset, We set $D$ to be the NYT training set,
and $C$ as to be the NYT development set.
\footnote{
Noting that there is no instance leakage in the following evaluation on development set:
we have remove it from our obtained clean set (line 10 of Algorithm \ref{alg:alg}).
}

\paragraph{ACE05-N} The ACE05-N dataset is a synthetic noisy dataset 
which adapted from ACE05 \cite{ace05}, a commonly used 
instance-level supervised dataset.
We first add the same amount of negative instances (with $\mathtt{None}$ relation label) as the annotated instances, 
and then mix additional noisy instances with different ratio,
which are flipped $\mathtt{None}$ instances. 
Detailed dataset specification could be found in the supplementary.

\paragraph{Settings} The settings and implementation details are 
in Appendix \ref{details}.
 We evaluate Precision, Recall, and F1 with micro-averaging in instance-level.

\subsection{Baselines}




\paragraph{ATT} \cite{lin-etal-2016-neural} is a classical bag-level denoising method
which tunes the attention weight of each instance in bags during training to alleviate the impact of noisy instances. 

\paragraph{RL} \cite{qin-etal-2018-robust} introduces reinforcement learning method to
train a instance selector that could tell the noisy instances from the distant supervised training set.

\paragraph{ARNOR} \cite{jia-etal-2019-arnor} embeds the relation pattern attention 
based on recurrent neural network into a bootstrapping framework. 

\paragraph{Confidence} We implement another baseline for fair comparison, which uses the trained model parameters $\hat{\theta}$ to select instances by confidence each iteration instead of influence function criteria. As a control, it also starts from a initial clean seed set.

\begin{table}[t]
    \begin{center}  
   \resizebox{\linewidth}{!}{
    \begin{tabular}{llcccccc}
    \toprule
    \multirow{2}*{\textbf{Encoder}} & \multirow{2}*{\textbf{Method}} & \multicolumn{3}{c}{\textbf{Dev}}& \multicolumn{3}{c}{\textbf{Test}}   \\ 
                          ~ & ~ & \textbf{Prec.} & \textbf{Rec.} & \textbf{F1}  & \textbf{Prec.} & \textbf{Rec.} & \textbf{F1}   \\   
    \midrule
    \multirow{5}*{\textbf{CNN}} & RL &42.50  &\textbf{71.62}  &53.34  &43.70  &72.34  &54.49 \\ 
    \cmidrule(lr){2-8}
    ~ & \texttt{Conf} &\textbf{83.41} &56.03 &67.03  &58.09  &58.09  &67.75\\   
    ~ & \texttt{Cr1} &81.33  & 43.94 &57.06  &73.49  &43.62  &54.75 \\ 
    ~ & \texttt{Cr2} &76.82 &61.54 &68.34 &\textbf{79.71} &60.48 &\textbf{68.78}\\
    ~ & \texttt{Cr2TS} &76.80 &62.10 &\textbf{68.69}  &75.36  &\textbf{60.52}  &67.13 \\ 
    \midrule
    \multirow{5}*{\textbf{PCNN}} & ATT &68.09 &47.49 &55.95 &67.31 &49.83 &57.27 \\ 
    \cmidrule(lr){2-8}
    ~ & \texttt{Conf} &82.18 &57.23 &67.47 &\textbf{80.15} &58.48 &67.62\\
    ~ & \texttt{Cr1} &78.38 &47.64 &59.26 &76.63 &52.19 &56.85 \\ 
    ~ & \texttt{Cr2}  &75.94 &\textbf{62.87} &68.78 &78.71 &\textbf{61.86} &\textbf{69.27}\\
    ~ & \texttt{Cr2TS} &\textbf{79.34} &61.14 &\textbf{69.06}  &79.60  &59.62  &68.17 \\ 
    \midrule
    \multirow{5}*{\textbf{BiLSTM}} & ARNOR &78.14$^*$  &59.82$^*$  &\textbf{67.77}$^*$  &\textbf{79.70}$^*$  &\textbf{62.30}$^*$  &\textbf{69.93}$^*$\\
    \cmidrule(lr){2-8}
    ~ & \texttt{Conf} &80.37 &55.82 &65.88 &79.46 &56.43 &65.99\\ 
    ~ & \texttt{Cr1} &\textbf{80.73} &55.28 &62.06 &69.28 &54.16 &60.79\\ 
    ~ & \texttt{Cr2} &72.39 &\textbf{60.67} &66.01 &72.04 &61.85 &66.56\\ 
    ~ & \texttt{Cr2TS} &77.38 &60.00 &67.59  &74.90  &58.65  &65.78 \\
    \bottomrule
     \end{tabular}
   }
    \caption{Comparison of our method and other baselines with different encoders. We denote \texttt{Conf}, \texttt{Cr1}, \texttt{Cr2}, \texttt{Cr2TS} as Confidence, Criterion 1, Criterion 2 and Criterion 2 with teacher-student update style.
    The code of ARNOR is not accessible now and we find it is hard to reproduce the reported performances.
    } 
    \label{main_result}
    \end{center}
\end{table}

\subsection{Main Result}
Table \ref{main_result} lists overall performances on NYT dataset
with different relation classification models
(recall that our approach is model-agnostic).
We compare the results of our method with several baselines.  
From the results, we find that,
\begin{itemize}[leftmargin=*]
  \setlength\itemsep{0em}
    \item Comparing with prior methods, both \texttt{Cr2} and \texttt{Cr2TS} achieve better or comparable performance with different encoders, 
  which suggests that our model-agnostic method could effectively prevent RE model from noise data.
    \item Both \texttt{Conf} and \texttt{Cr2} use the dev data as a reference,
  while \texttt{Cr2} achieves superior performance.
  Thus, our method is a better strategy which makes use of limited clean set.
  We credit it to that the influence function could select better instances under the criterion 2.
    \item The results show that \texttt{Cr1} performs much worse than \texttt{Cr2} on both dev and test set.
    As we mentioned before, \texttt{Cr1} uses dirty set as training set, leading to unreliable influence.
   \item The performance of \texttt{Cr2TS} is worse than that of \texttt{Cr2}, which shows the teacher-student mechanism has no advantage on this dataset.
    We guess it's relative to the noise ratio on dirty set, and further discuss in next section.
\end{itemize}

\section{Analysis and Discussion}

\paragraph{Validating influence function.}
The calculation of influence function is the key step of our method. 
Here we show the high correlation between the  real influence (calculated by leave-one-out retraining) 
and estimated influence.
From the experimental results (see Appendix \ref{influence}), we find that the correlation among high influential instances is 0.79,
and 0.65 in all instances.
The high correlation validate that
influence function is reliable in perform instance perturbation analyses.



\begin{figure*}[t]
    \centering
    \includegraphics[width=14.1cm,height=4.7cm]{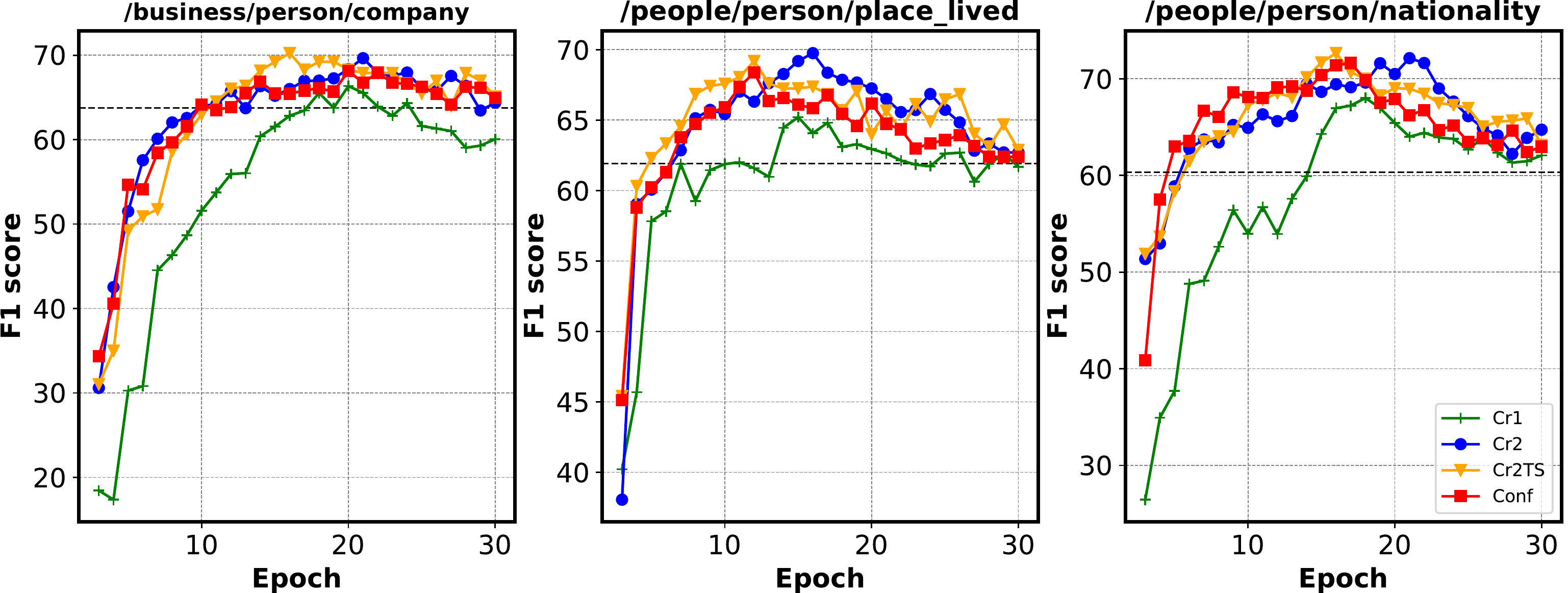}
    \caption{Bootstrapping result. Here we take 3 NYT relations as examples, we present  performance change on dev  during the bootstrapping. The black dash lines are the performance without denoising.}\label{BS}
  \end{figure*}

\begin{figure*}[t]
  \centering
  \includegraphics[width=14.1cm,height=4.7cm]{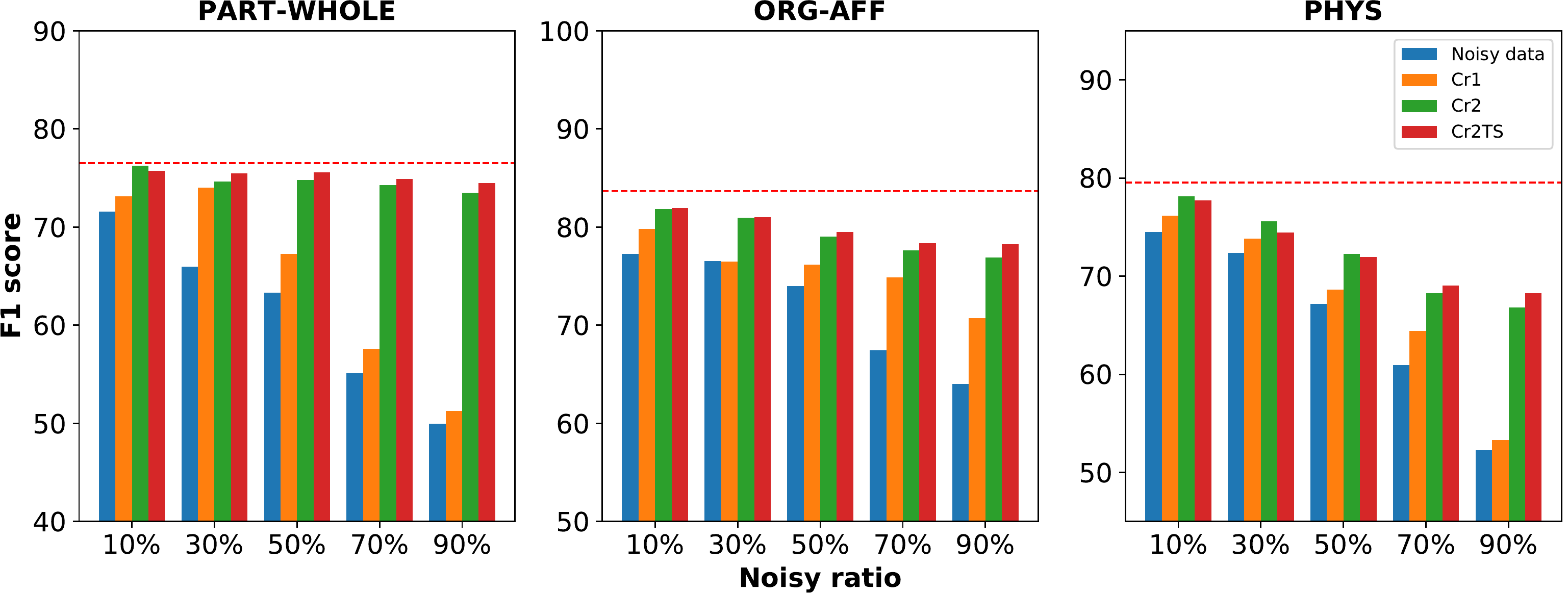}
  \caption{The denoising ability of our method with different noise ratio. Here we take three relation types from ACE05-N as examples. We evaluate their performances with the different ratio of noisy instances that range from 10\% to 90\% for each relation type. The red dash line indicates the performance without any noisy.}\label{NR}
\end{figure*}

\paragraph{Bootstrapping process in detail.}
In this section, we study the performance change of four selecting strategies during the bootstrapping procedure, as show in Figure \ref{BS}. 
\begin{itemize}[leftmargin=*]
\setlength\itemsep{0em}
\item The performance curves of four strategies are quite similar,
which gradually rise to the peak at the beginning and then fall to the line of original noisy data.
We think the main reason for this phenomenon is that these strategies add more clean instances into $D_t^c$ in the early epochs, and inevitably select more and more noisy instances in the later training epochs.
\item The curve of \texttt{Cr2} and \texttt{Cr2TS} is  higher than
\texttt{Conf}, which suggests that the effectiveness of criterion 2. 
As expected, \texttt{Cr1} fails in the later period, which is even inferior to training with original noisy data.
\end{itemize}

\paragraph{The impact of noise ratio.}
We conduct experiments with different ratio of noise data to verify the denoising ability of our method (Figure \ref{NR}).
\begin{itemize}[leftmargin=*]
\setlength\itemsep{0em}

\item Even the noise ratio is extreme high (90\%), the \texttt{Cr2} and \texttt{Cr2TS} is still stable.
We think that our methods take the most of the clean set to distinguish clean instances from the dirty set, so the damage from the noise instances in dirty set is quite slight.

\item The \texttt{Cr1} would crash when the noise ratio beyond a certain level, as we analysis before, 
the higher noise ratio the train set has, the wider gap between estimated influence and the real influence of noisy instances.

\item It is worth noting that \texttt{Cr2TS} would be better than \texttt{Cr2}
in the case of the high noise ratio, which shows the effectiveness of teacher-student mechanism.
The teacher-student mechanism has the advantage to help lower the lasting impact of misclassified instances in the previous iteration. 
\end{itemize}


\paragraph{The impact of initial clean set size.}
\begin{table}[t]
\small
  \begin{center}  
  \begin{tabular}{c|cccc}
  \toprule
  \textbf{Methods} &  \textbf{10} &  \textbf{30} &  \textbf{50} & \textbf{ALL}  \\   
  \midrule
        Conf &31.43 &34.28 &35.89 &36.95   \\       
        Cr1 &27.93 &28.22  &28.63 &30.73   \\  
        Cr2 &35.1 &36.23 &37.42 &37.87 \\
        Cr2TS &36.52 &37.82 &38.04 &38.69  \\
  \bottomrule       
  \end{tabular}
  \caption{We conduct our experiment on ACE05-N with 50\% noise ratio to study the 
  impact of initial clean set size.
  Note that the F1-score is 27.50 without any selecting strategy.
  Columns represents the result of  using different number of instances as initial clean set.
  For each relation, we try to use 10, 30, 50 and all dev set as the intial clean set.}   
  \label{CS}
  \end{center}
  \vspace{-0.5cm}
\end{table}

Our method starts with an initial clean set, 
so we study the impact of the set size in Table \ref{CS}.
\begin{itemize}[leftmargin=*]
\setlength\itemsep{0em}
\item In general, the performance goes down with the number of clean instances decreasing. 
That is reasonable for that the limited clean seed set would suppress the methods to find more true positive instances.
\item The performance of \texttt{Conf} drops sharply with few initial clean instances (10 instances).
We guess that the method only considering confidence of instances is easily trapped into
the limited clean set and hard to detect more clean instances.
\item Both \texttt{Cr2} and \texttt{Cr2TS} show better robustness even the 
 the size is extreme small. 
 We believe that the key factor is the influence function, 
 which considers more than confidence and is more practical to extend the scale of
 clean instances from a small start. 
\end{itemize}

\paragraph{Stratification of instances.}
Table \ref{CaseStudy} presents a stratification of instances in the noisy dataset,
 which is our source of inspiration. 
There are three layers sorted by the score with Criterion 2.\footnote{We just take one reference instance as example, rather than the average of all reference instances in criterion 2.}
The first layer contains instances with large positive score, 
which usually have a similar syntactic and semantic structure with the reference instance.
These instances are true positive instances, and our method select them in every iteration.
The second layer is made up of instances with score around zero.
These instances are usually hard to tell whether they are noisy or not.
The true positive instances in this layer could be discovered by extending clean set with bootstrapping.
The last layer is formed by instances with large negative scores, 
which are quite different from the reference instance.
Some of these instances are indeed noise, 
while some are still true positive instances but just not be supported by this reference instance.
These true positive instances would be supported by other reference instances in clean set which selected by the average score in Criterion 2.

\begin{table}
    \begin{center}
        \resizebox{\linewidth}{!}
        {
        \begin{tabular}{|p{1.4cm}|p{4.5cm}|c|c|}
        \hline
         \multicolumn{4}{|p{8cm}|}{\textbf{Reference}:\underline{Jeffrey Katzenberg},
             chief executive of \underline{DreamWorks Animation}, said\dots} \\\hline
        &\textbf{Sentence}& \textbf{TP/FP}&\textbf{Score} \\ \hline
        \multirow{5}{*}{\shortstack{Obvious \\ \\ TP}} &\dots \underline{Richard C.Notebaert}, the cheif excutive of \underline{Qwest}\dots & \multirow{2}{*}{TP} & \multirow{2}{*}{3.46e-3} \\ \cline{2-4}
                           &\dots and \underline{Bruce Wassertein},the chairman and cheif excutive of \underline{Lazard} & \multirow{3}{*}{TP} & \multirow{3}{*}{1.73e-3}\\ \hline
        \multirow{6}{*}{\shortstack{Hard \\  \\ instances}} &\dots last October, \underline{Ray Ozzie}, chief technical officer, who joined \underline{Microsoft} last year \dots & \multirow{3}{*}{TP} & \multirow{3}{*}{3.39e-5} \\ \cline{2-4}
                           &\dots \underline{Richard C.Noteaert}, the company's chief executive, said \underline{Qwest} spent..  & \multirow{3}{*}{FP} &  \multirow{3}{*}{4.91e-5}\\ \hline
        \multirow{8}{*}{\shortstack{Potential \\ \\ FP}}
                    &\underline{Eric Foner} is the De Witt Clinton professor of history at \underline{Columbia University} and the author\dots & \multirow{4}{*}{TP} & \multirow{4}{*}{-2.36e-3}\\ \cline{2-4}
                     &As \underline{Bruce Wasserstein} left St. Regis Hotel in Manhattan on Tuesday afternoon after presenting \underline{Lazard}'s plan \dots & \multirow{4}{*}{FP} & \multirow{4}{*}{-4.37e-3} \\ \hline
        \end{tabular}
     }
        \caption{An example of layered phenomenon of instances in the noisy dataset. We group instances by their scores calculated by \texttt{Cr2}.}
        \label{CaseStudy}
    \end{center}
    \vspace{-0.6cm}
\end{table}
\section{Related Work}
We focus on distant supervision relation extraction via influence function in this paper.
For relation extraction, various neural networks like CNN \cite{zeng-etal-2014-relation, zeng-etal-2015-distant}, 
RNN \cite{zhang-etal-2015-bidirectional} and Tree-GRU \cite{AAAI1816362}.
Distant supervision provides a method to automatically label massive training data \cite{mintz-etal-2009-distant},
meanwhile, bringing excessive wrong label instances, so called noise,  which stems the training.

To solve the noisy problem, people first take a multi-instance learning methods \cite{surdeanu-etal-2012-multi,lin-etal-2016-neural,ye-ling-2019-distant},
which puts instances with same entity pair into bags, to alleviate the impact of noisy instances.
Then, to make training process closer to real-world application, people focus on instance-level denoising method. 
An instance-selector is utilized to pick out trustable instance,
which is trained by reinforcement learning \cite{qin-etal-2018-robust, feng2018reinforcement} and 
adversarial learning \cite{qin-etal-2018-dsgan}. 
The bootstrapping framework \cite{jia-etal-2019-arnor, Li-2020} 
is also utilized to  promote the ability of classification model gradually from a small seed.
For influence function, which is commonly used in robust statistics \cite{cook1982},
\citet{koh2017understanding} introduce it in machine learning area. 
As a technology that is aiming to analyse the every training points' influence on model prediction,
influence function is widely applied. \citet{ren2020unlabeled} apply influence function on weighting unlabeled data
to promote semi-supervised learning. \citet{ActiveLearning} utilize influence function to designed an efficient strategy
for active learning.

\section{Conclusion}
In this paper, we propose a model-agnostic denoise method for distant supervision relation extraction. 
We start from training a relation classifier on the clean set and propose a new criterion to select good instances from the noisy data.
We leverage the criterion in a bootstrapping learning to extent the clean set iteratively. Further, we propose a teacher-student to control the update.
 Our method has a strong performance on NYT dataset and shows robustness under the high noise ratio circumstance or very limited size of initial clean set on the synthetic ACE05-N dataset.

\section{Acknowledgments}
We thank all anonymous reviewers for their constructive and helpful feedback. 
The corresponding author is Yuanbin Wu. This research was supported by NSFC (62076097) 
and a joint research fund from AI Innovation Center, Midea Group.

\bibliography{acl2022}
\appendix
\clearpage
\twocolumn[
\begin{@twocolumnfalse}
    \section*{ \centering{ Supplementary Materials for \\ \emph{Few Clean Instances Help Denoising Distant Supervision\\[30pt]}}}
\end{@twocolumnfalse}
]

\section{Proof of Lemma 1}
\label{proofSection}

\begin{proof}
  Following the same first order Taylor approximation as influence function,
  up to $o(n^{-1})$, we can get the update of parameters after replacing $z$ with $z'$ 
  (see Equation 3 of \cite{koh2017understanding}),
  \begin{IEEEeqnarray*}{rl}
      \hat{\theta}_{z,z'} - \hat{\theta} &= 
      n^{-1} \left(\mathcal{I}_{up,params}(z) - \mathcal{I}_{up,params}(z') \right) \\
      & = n^{-1} H_{\hat{\theta}}^{-1}\left( 
  \nabla_{\theta}\mathcal{L}(z,\hat{\theta}) - 
  \nabla_{\theta}\mathcal{L}(z',\hat{\theta}) \right).
  \end{IEEEeqnarray*}
  Since the set $D$ is finite, we denote
    \begin{IEEEeqnarray*}{c}
        c_1 \triangleq \arg{\min}_{(z, z')} \frac{|\mathcal{L}(z, \hat{\theta}_{z,z'})-\mathcal{L}(z, \hat{\theta})|}
        {\|\hat{\theta}_{z,z'} - \hat{\theta}\|},
    \end{IEEEeqnarray*}
  then
  \begin{IEEEeqnarray*}{rl}
      &|L(z_d, \hat{\theta}_{z,z'}) - L(z_d, \hat{\theta})|
       \geq c_1 \|\hat{\theta}_{z,z'} - \hat{\theta}\| \\
       = & c_1 \|H_{\hat{\theta}}^{-1}\left( 
        \nabla_{\theta}\mathcal{L}(z,\hat{\theta}) - 
        \nabla_{\theta}\mathcal{L}(z',\hat{\theta}) \right)\|\\
    \geq & c_1 (n\sigma)^{-1}\|
        \nabla_{\theta}\mathcal{L}(z,\hat{\theta}) - 
        \nabla_{\theta}\mathcal{L}(z',\hat{\theta})
        \|,
  \end{IEEEeqnarray*}
  where $\sigma$ is the maximum singular value of the Hessian.
  Regarding the log-likelihood loss, $\nabla_{\theta}\mathcal{L}(z,\hat{\theta})=
  \nabla_\theta\log Z - \nabla_\theta \hat{w}_y^\intercal h(x, \hat{\varphi})$, 
  and the transpose of the second term is 
  \begin{equation*}
      \bordermatrix{
       & & w_y & \varphi\cr 
       & \mathbf{0}, & h^\intercal(x, \hat{\varphi}), 
       &  \hat{w}_y^\intercal \nabla_{\varphi} h(x, \hat{\varphi})\cr
     }.
  \end{equation*}

  Hence, 
  \begin{IEEEeqnarray*}{rl}
  & \|
    \nabla_{\theta}\mathcal{L}(z,\hat{\theta}) - 
    \nabla_{\theta}\mathcal{L}(z',\hat{\theta})
  \| \\ 
  = & \sqrt{2\|h(x, \varphi)\|^2 + \|\nabla_{\varphi} h(x, \varphi)^\intercal (w_y - w_{y'})\|^2} \\
  \geq  & \|h(x, \varphi)\| + \frac{\tau_{x}}{\sqrt{2}}\|w_y - w_{y'}\|.
  \end{IEEEeqnarray*} 
  Let $c = c_1(\sigma\sqrt{2})^{-1}$, we get the lower bound.

\end{proof}

\section{Implementation details}\label{details}
For basic CNN model,
the window size of the convolution layer is set to 3 and the number of the filter is set to 230.
In bootstrapping procedure, 
the position embedding dimension of CNN is set to 1 and the word embedding  is initialized with 
100 dimensional pre-trained  glove embedding \cite{DBLP:conf/emnlp/PenningtonSM14},\footnote{Download from \url{https://nlp.stanford.edu/projects/glove/}.}
which is for making IF focuses more on semantic information. 
In training procedure,
for fair comparison, 
the position embedding dimension of all models is set to 5,
the word embedding dimension is set to 100 with random initialization 
and an entity type embedding. 
And for the PCNN model we have the same hype-parameters with \citet{zeng-etal-2015-distant}.
For majority vote,we set $k=3$.
In selection step of bootstrap, we at most select $n=\frac{1}{10} |{D}_{t}|$ instances.
For teacher-student, the $\alpha$ is set to 1 and $\beta$ is set to 0.9.
To avoid the model just memorizes the entity pairs, we mask the entity words in both bootstrapping and training.

\section{Detailed statistics on NYT dataset}

\begin{table}
    \centering
    \begin{tabular}{cccc}
    \toprule
    \textbf{NYT} &\textbf{Training} & \textbf{Dev} &\textbf{Test} \\ 
    \midrule
    \# Sentence &233038 & 1596 & 1596  \\ 
    \# Instance &367596 & 4567 & 4484   \\ 
    \# Positive instances &106653 & 975 & 1050   \\ 
    \bottomrule
\end{tabular}
\caption{Statistics on NYT dataset.}
\label{statNYT}
\end{table}

\begin{table}
    \centering
    \resizebox{\linewidth}{!}{
    \begin{tabular}{lccc}
    \toprule
    NYT &Training &Dev &Test \\ 
    \midrule
    \# /people/person/place\_lived & 7197 & 198 & 185  \\ 
    \# /location/location/contains &51766 & 479 & 611   \\ 
    \# /people/person/nationality  &8079 & 117 & 91   \\ 
    \# /business/person/company &5595 & 105 &  113 \\
    \# /people/person/children  &506 & 6 &  11\\
    \# /people/dec\dots/place\_of\_death &1936 & 8 &  14\\
    \# /location/country/capital &7690 & 14 &  15\\
    \# /business/company/founders &800 & 10 &  6\\
    \# /people/person/place\_of\_birth &3173 & 13 &  15\\
    \# /location/nei\dots/nei\dots\_of &5553 & 6 &  7\\
    \bottomrule
\end{tabular}
}
\caption{10 relations on dev set. Our methods take the dev set as
clean set to denoise.}\label{statNYT2}
\end{table}

Overall statistics on NYT datset are shown in 
Table \ref{statNYT}. And the detailed statistics of each relations on dev set are in Table \ref{statNYT2}.

\section{Synthetic dataset ACE05-N}
\begin{table}
    \centering
    \resizebox{\linewidth}{!}{
    \begin{tabular}{lccc}
    \toprule
    ACE05-N &Training &Dev &Test \\ 
    \midrule
    \# GEN-AFF &290 & 73 &  55 \\ 
    \# ORG-AFF &857 & 204 & 203   \\ 
    \# PER-SOC &279 & 57 & 45 \\ 
    \# PHYS    &549 & 164 &  123 \\
    \# PART-WHOLE &393 & 81 & 86\\
    \# ART     &275    & 52 & 85\\
    \# NA   &116572    &27597 &24363\\ 
    \bottomrule
\end{tabular}
}
\caption{The statistics on ACE05.}\label{statACE}
\end{table}

Table \ref{statACE} shows the statistics on ACE05.
Next we show how to construct our synthetic dataset ACE05-N,
which takes two steps: adding NA and adding noise.
The entity pair in a sentence that does not express any positive relation would be 
considered as ``NA”. 
The NA instances play two roles in our synthetic dataset:
the negative instances and noisy.

\noindent \textbf{Adding negative instances} In DSRE, 
the noise come from the wrong-labelled negative instances.
So the true negative instances are necessary for evaluating the denoise ability.
In our synthetic dataset, we reconstruct the training, dev and test of each relation
by adding the NA. The size of NA is same as the original size of training, dev and test.
After that, the dataset would be changed to Table \ref{statACE2}.
\begin{table}
    \centering
    \resizebox{\linewidth}{!}{
    \begin{tabular}{lccc}
    \toprule
    NYT &Training &Dev &Test \\ 
    \midrule
    \# GEN-AFF &580 & 146 &  110 \\ 
    \# ORG-AFF &1714 & 408 & 406   \\ 
    \# PER-SOC &556 & 104 & 90 \\ 
    \# PHYS    &1098 & 328 &  246 \\
    \# PART-WHOLE &786 & 162 & 172\\
    \# ART     &550   & 104 & 170\\
    \bottomrule
\end{tabular}
}
\caption{The statistics of ACE05 after adding NA.}\label{statACE2}
\end{table}

\noindent \textbf{Adding noisy} An instance that don't express relation $r$ but labelled with $r$ is a noise instance
to the relation $r$. So the intentionally made noise is  relabelling a NA instance to a positive label.
In the experiment, we manually put a certain number of noise instance to poison the dataset.
If we poison 
a train set  with 50\% noise, we mean that after poisoning,
the noise ratio of this train set is 50\%.
For example, if the train set  of \emph{GEN-AFF} in Table \ref{statACE2} is poisoned with 50\% noise,
it would be put with 580 noise instances.

\section{Validating influence function}\label{influence}
\begin{figure}[t]
    \centering
    \includegraphics[width=5cm,height=4.5cm]{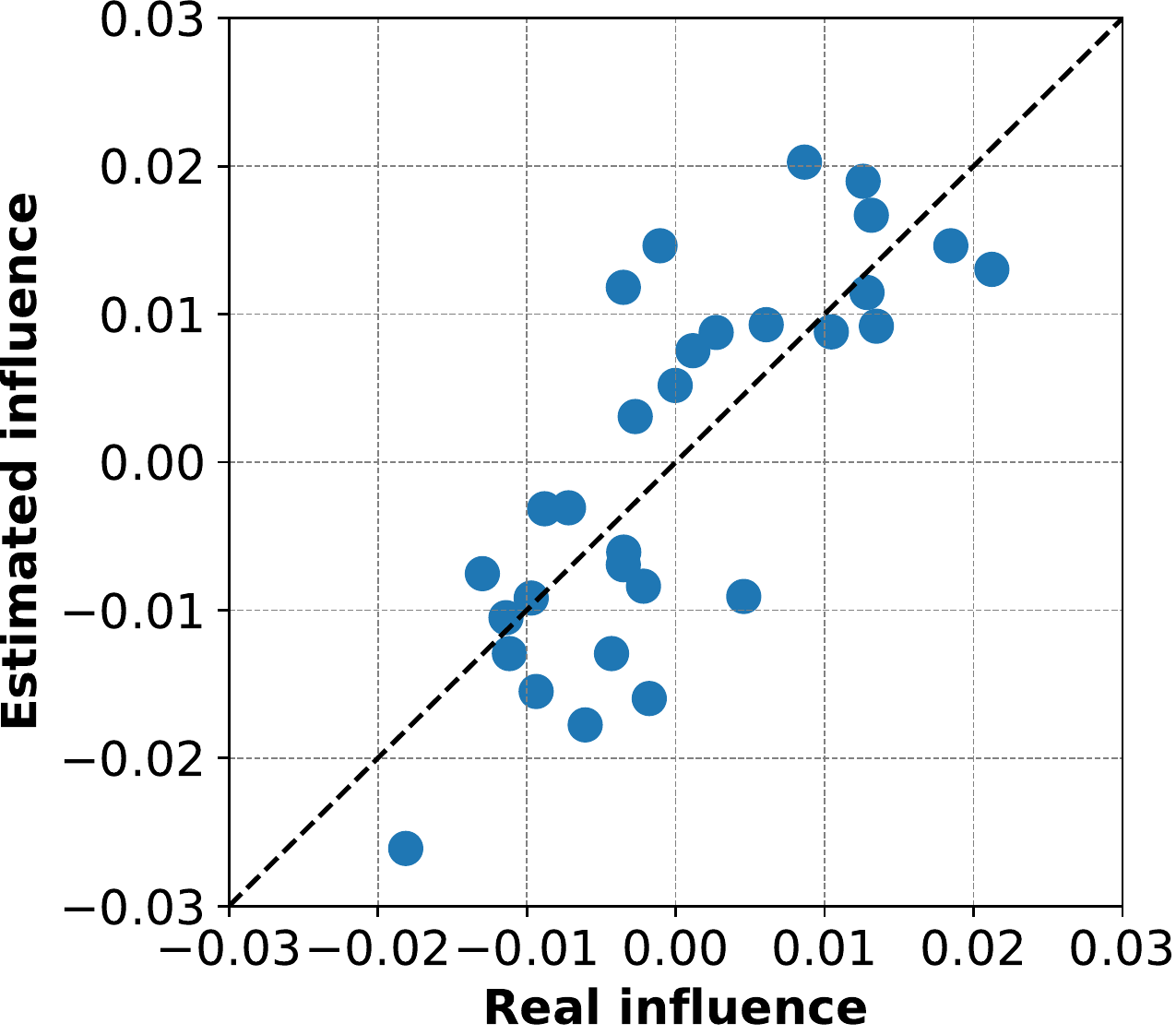}
    \caption{The correlation between estimated change of loss and real change of loss.
    We use a training set with 500 instances included two relation types and arbitrarily pick four instances as testing instances to validate computation of influence function.
    The picture shows 40 most influential points  
    with their real difference in loss
    (obtained by 500 steps leave-one-out retraining).
}\label{Cor}
\end{figure}
The calculation of influence function is the key step of our method. 
Here we show the high correlation between the  real influence (calculated by leave-one-out retraining) 
and estimated influence.
From the experimental results (Figure \ref{Cor}), we find that the correlation among high influential instances is 0.79,
and 0.65 in all instances.
The high correlation validate that
influence function is reliable in perform instance perturbation analyses.

\end{document}